# Word Segmentation for Asian Languages: Chinese, Korean, and Japanese

**Matthew Rho**
College of Computer Science
Georgia Institute of Technology
Atlanta, GA, 30308
mrho3@gatech.edu

**Yexin Tian**
College of Computer Science
Georgia Institute of Technology
Atlanta, GA, 30308
yexintian@gatech.edu

**Qin Chen**
College of Computer Science
Georgia Institute of Technology
Atlanta, GA, 30308
qchen352@gatech.edu

## Abstract

We provide a detailed overview of various approaches to word segmentation of Asian Languages, specifically Chinese, Korean, and Japanese languages. For each language, approaches to deal with word segmentation differs. We also include our analysis about certain advantages and disadvantages to each method. In addition, there is room for future work in this field.

# 1 Introduction

## 1.1 Motivation of Word Segmentation

Breaking a long sentence into several meaningful units is usually the first step for humans to understand what this sentence tells us and it is also the first step for the machine to process human languages. Thus, word segmentation is important and is influential in many fields including developing text processing applications, such as Information Extraction, Document Summarization, Machine Translation (MT), Natural Language Processing, Information Retrieval, Language Modeling, and Speech Recognition.(15) Word segmentation is often a vital task of language processing.

In addition, the reason why word segmentation is significant in the field of Natural Language Processing is because it is the initial step for most higher level natural language processing tasks, such as part-of-speech tagging and parsing. In addition, for languages that are space-delimited such as English or Russian, these languages are being segmented differently as opposed to those that don't have explicit word boundary delimiters, such as Chinese and Japanese. There is a common goal for this task, which is to have a near-perfect word segmentation system, which can still perform reasonably with no or minimum language-specific adaptations (9).

## 1.2 Introduction of Word Segmentation of Asian Languages

However, unlike English, many East Asian languages (e.g. Korean, Japanese, and Chinese) lack a trivial word segmentation process. This can lead to difficulties for downstream NLP applications (English) I love Natural Language Processing.
(Chinese) 我喜欢自然语言处理。
(Korean) 저는 자연어 처리를 좋아합니다.
(Japanese) 私は自然言語処理が大好きです。

Figure 1: The sentence "I love Natural Language Processing" in four different languages: English, Chinese, Korean, and Japanese.

like machine translation. Thus, it is crucial to perform accurate word segmentation for these languages. For example, Figure 1 shows how the sentence "I love Natural Language Processing" is being represented in the four different languages: English, Chinese, Korean, and Japanese. We can clearly tell that these three Asian languages, unlike English, do not have white spaces to separate each word. This is one of the major reasons why word segmentation of these Asian languages are more difficult.

For a long time, CKJ (Chinese, Korean, Japanese) Word Segmentation has been widely concerned in the research of NLP. No matter before the rise of deep learning or since the rise of deep learning, the research on CKJ (Chinese, Korean, Japanese) Word Segmentation has never stopped. Although in terms of form, "character" is the smallest combination of sound and meaning, in modern CKJ languages, "word" has the function of expressing complete semantics, and most "words" are combinations of multiple "characters". Therefore, CKJ word segmentation has become the first step of many CKJ NLP tasks.

In this survey, we will go over three different methods of each language in order from least accurate to most accurate, and discuss the advantages and disadvantages of each of them. To be more specific, the survey will organized as follows. Section 2, 3, and 4 presents Korean, Chinese, and Japanese respectively. Section 5 outlines related works and Section 6 includes our conclusions as well as future word. Section 7 briefly summarized our team responsibility.

## 2 Korean

Korean, in contrast to Chinese, has various morphological structures and modal expressions that make word segmentation challenging. Main challenges arise since there is a high level of sparsity in Korean phrases and sentences.

### 2.1 Dependency Grammar Parsing

In order to tackle the issue of sparsity of Korean, one study used a dependency grammar for the parsing task. This dependency grammar is similar to the dependency grammar that is used to parse English. However, there are drawbacks to using this same dependency grammar since there are often many incorrect dependency relations that occur during parsing, especially in Korean. In order to tackle this potential problem, they introduced various dependency constraints such as having dependents precede heads, and that dependency grammars cannot cross, and that every verb isn't allowed to have more than one subject and object.

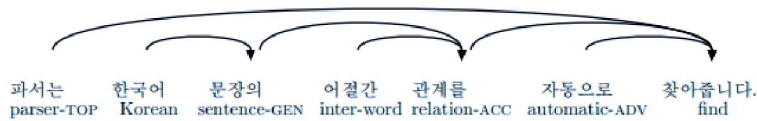

There are different ways that dependency parsing is used, for instance, one study used statistical dependency parsing, and another study used a more hybrid form of dependency parsing, using valency information and structural preference rule with statistical information from the KAIST Treebank for resolving ambiguities in dependency parsing (Chung, 2004) (2).

The statistical dependency parsing method is the main focus of Chung's study. The parser selects a tree among all parses of a Korean sentence. And the parser uses some simple constraints to reflect the syntactic characteristic of Korean, to prevent over-generation of dependency arcs, and to promote efficiency and accuracy.

The parameters for the parsing model is estimated by using the lexical dependency probability, which is a conditional probability measure of the head given multiple parts of the sentence. To cope with the sparse data problem, linear interpolation is used to estimate the parameter. In addition, there is a parameter for modification distance probability with their linear interpolations for estimating the lexical dependency probability being estimated with back-off smoothing.

All experiments were done using the dependency-tagged KAIST Treebank from KAIST Language Resources. The Treebank consists of 31,080 sentences and stored in 100 files. Chung used 91 files (27,694 sentences) as the training data and the 9 files (3,386 sentences) as the heldout testing data. The average lengths of the sentences are 12.2 words and 12.45 words for training and testing set, respectively.

|  | Measures | $K=2$ ($\{1, \text{long}\}$) | $K=3$ ($\{1,2, \text{long}\}$) | $K=4$ ($\{1,2,3, \text{long}\}$) |
|---|---|---|---|---|
| Training Set | Arc $F_1$ | 0.9897 | 0.9842 | 0.9804 |
|  | ExactMatch | 0.9023 | 0.8631 | 0.8333 |
| Testing Set | Arc $F_1$ | 0.8581 | 0.8668 | 0.8674 |
|  | ExactMatch | 0.3452 | 0.3467 | 0.3405 |

Figure 2: Parsing Results on the training and testing data

As shown in Figure 2, the parser achieved a relatively high Arc-based $F_1$ measure (arc-based meaning that the measure is based on number of correct arcs and number of total arcs, with same calculation of precision and recall and $F_1$). (The ExactMatch is the formula of number of correctly analyzed sentences over the number of sentences in test data, and the K scores represent the distance of the root word from the words that are 1 to K words from the root word as shown in the Figure 3 as reference)

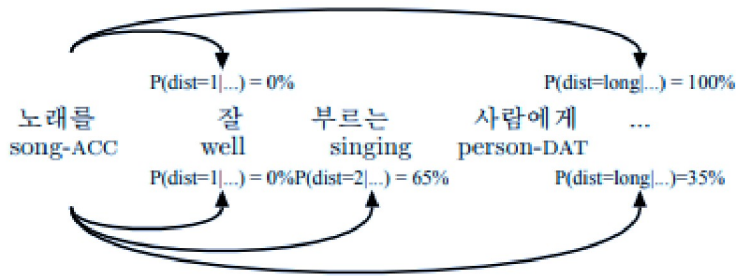

Figure 3: Dependency parse sample

As shown in Figure 4, there were 4 different models that were compared to the model that Chung proposed. Model 1 is the baseline lexical dependency probability model. Model 2 is the lexical dependency probability model that distinguishes dependencies between adjacent words from other lexical dependencies. Model 3 is the lexical dependency probability model combined with triplet/quadruplet head candidate decision model. Model 4 is the parsing model based on bigram lexical dependency and distance between depending words.

All 4 of these models were trained and tested on the same data that was used for the proposed model. As shown in the figure, the proposed model does not perform well in the training set, compared to the other models. However, the proposed model performed better than all of the 4 models in the testing set.

|  | Measures | Model1 | Model2 | Model3 | Model4 | Proposed |
|---|---|---|---|---|---|---|
| Training Set | Arc $F_1$ score | 0.9937 | 0.9953 | 0.9443 | 0.9962 | 0.9804 |
|  | Exact Matching | 0.9476 | 0.9600 | 0.6444 | 0.9665 | 0.8336 |
| Testing Set | Arc $F_1$ score | 0.7846 | 0.8302 | 0.8321 | 0.8370 | 0.8674 |
|  | Exact Matching | 0.2451 | 0.2548 | 0.2826 | 0.2569 | 0.3405 |

Figure 4: Results comparison

## 2.2 Transition-based Bi-LSTM parser

While the previous section focused on the character level parsing through the Korean text, one study took a step further and focused on the "jamo" letter components of each character. A "jamo" letter is the phonetic symbols that make a Korean character, such as ㄱ, ㄴ, ㄷ, ㅇ, and etc (known as the Korean phonetic alphabet). In order to parse at the jamo letter level, the study used a transition-based Bi-LSTM parser, which parses through each jamo letter using Unicode Decomposition. Here is a sample jamo letter parse tree based on the sentence "I love Natural Language Processing" (used in Figure 1) in Korean (underlined fragments are selected and parsed in the three separate parse trees below):

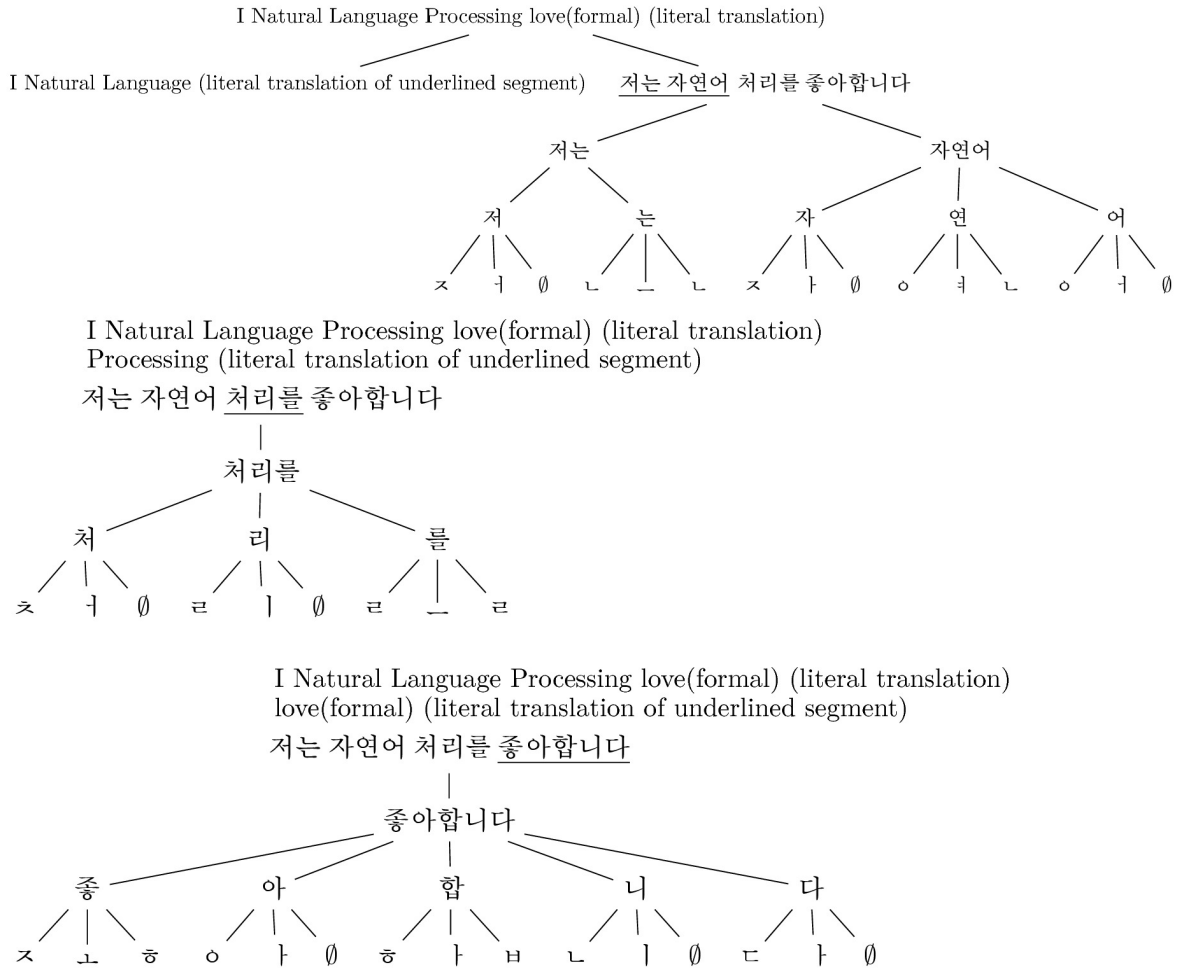

(Parsing diagrams are based on the parsing diagram in Stratos' study (Stratos, 2004) (12))

A brief description of the BiLSTM method that Stratos used:
Firstly, Stratos constructed the jamo architecture to be plugged in to the BiLSTM model. He considered three parts to Korean characters, the head, the vowel, and the tail. In order to pull the jamo letters from the characters, he used unicode decomposition.

Unicode decomposition method:
The process of extracting jamo letters dynamically from any possible Korean character is explained as follows. For any Korean character $c \in C$ with Unicode value $U(c)$, let $U\!\!\!/(c) = U(c) - 44032$ and $T(c) = U\!\!\!/(c)$ mod 28. Stratos then calculates the Unicode values of the head, vowel, and tail consonant as follows:

$U(c_{head}) = 1 + \left\lfloor \frac{U\!\!\!/(c)}{588} \right\rfloor + 0x10\text{ff}$

$U(c_{vowel}) = 1 + \left\lceil \frac{(U\!\!\!/(c) - T(c)) mod 588}{28} \right\rceil + 0x1160$

$U(c_{tail}) = 1 + T(c) + 0x11a7$

where $c_t$ is set to $\emptyset$ if $T(c_t) = 0$
(0x10ff and 0x1160 are unicode values)

After he pulled the jamo letters, he created a new representation of the letters by using the tanh function as follows:

$h^c = tanh(U^J e^{c_h} + V^J e^{c_v} + W^J e^{c_t} + b^J)$

(note: $U$, $V$, and $W$ are matrices of jamo letters, $e^{c_h}$, $e^{c_v}$, and $e_{c_t}$ are jamo letter head, vowel, and tail consonant embeddings respectively, and $b^J$ is an array of all possible 51 jamo letters)

This new representation is then concatenated with a character-level lookup embedding, and the result of the concatenation is then fed to an LSTM to produce a word representation. The LSTM provides a mapping by taking an input vector, x, and a state vector, h, to compute a new state vector $h'$ ($h' = \phi(x, h)$).

Here is a detailed list of the parameters in this layer of the LSTM:

- Character Embedding $e^c \in R^{d'}$ for each $c \in C$
    ($d'$ is the new dimension and c is a character)

- Forward LSTM mapping $\phi^f$: $\mathbb{R}^{d+d'} \times \mathbb{R}^d \to \mathbb{R}^d$
    (dimension mapping from $d + d'$th dimension x $d$th dimension to $d$th dimension)

- Backward LSTM mapping $\phi^b$: $\mathbb{R}^{d+d'} \times \mathbb{R}^d \to \mathbb{R}^d$
    (dimension mapping from $d + d'$th dimension x $d$th dimension to $d$th dimension)

- $U^C \in \mathbb{R}^{dx2d}$ and $b^C \in \mathbb{R}^d$
    ($U^C$ is the character matrix and $b^C$ is the character array)

Now, given a word $w \in W$ and its character sequence $c_1...c_m \in C$, Stratos computes:

$f_i^c = \phi^f(\begin{bmatrix} h^{c_i} \\ e^{c_i} \end{bmatrix}, f_{i-1}^c) \quad \forall i = 1, \ldots, m$

(i is the index of the characters in order of the sequence of characters $C$)

$b_i^c = \phi^b(\begin{bmatrix} h^{c_i} \\ e^{c_i} \end{bmatrix}, b_{i+1}^c) \quad \forall i = m, \ldots, 1$

(i is the index of the characters in order of the sequence of characters $C$)

And then, Stratos induced a new representation of $w$ using the tanh function as follows:

$$h^w = tanh(U^C \begin{bmatrix} f_m^c \\ b_1^c \end{bmatrix} + b^C)$$

(*tanh* of the matrix multiplication of the character matrix $U^C$ and the vector $\begin{bmatrix} f_m^c \\ b_1^c \end{bmatrix}$ and addition with the array of characters, $b^C$)

This new representation is then concatenated with a word-level lookup embedding and the result is fed into a BiLSTM network.

Here is a detailed list of the parameters in this layer of the LSTM:

- Word Embedding $e^w \in \mathbb{R}^{d_W}$ for each $w \in W$
    ($d_W$ is the dimension associated with the word sequence (sentence) $W$ and $w$ is a word)

- Two-layer BiLSTM $\Phi$ that creates a mapping: $h_1...h_n \in \mathbb{R}^{d+d_W} \to z_1...z_n \in \mathbb{R}^{d^*}$

- A greedy feed-forward network for predicting transitions based on words that are active in the system

Finally, given a sequence of words (sentence) $w_1...w_n \in W$, the final $d^*$ dimensional word representations are given:

$$(z_1...z_n) = \Phi(\begin{bmatrix} h^{w_1} \\ e^{w_1} \end{bmatrix} ... \begin{bmatrix} h^{w_n} \\ e^{w_n} \end{bmatrix})$$

With the word representations, $(z_1...z_n)$, and greedy feed-forward network, the BiLSTM model is then trained end-to-end by optimizing a max-margin objective, previously used by Kiperwasser and Goldberg (Kiperwasser and Goldberg, 2016).

With the embedding dimension of jamos, $d$, characters, $d'$, and words, $d_W$, being set to zero, Stratos configured the BiLSTM network to any combination of the input units.

For the implementation of the BiLSTM network, Stratos used a publicly available Korean treebank in the universal treebank 2.0 (Mcdonald et al., 2013).

Some benefits behind the use of jamo letters in our parsing method is that in the Korean language, there are certain jamo letters that can give implications of the tense of the word that the jamo letters are modifying (12). In addition, jamo letters also indicate the sound of the character which would help with speech processing (12). Although speech processing isn't the main focus of this survey, this phonetic element would help with lexical correlation (Stratos, 2004) (12).

In the results in Figure 2, Stratos' parser, KoreanNet, outperforms other parsers that use a combination of word, character, and jamo letter features, even with a small set of jamo letters with a Test UAS of 95.17 and a Test LAS of 92.31, the highest Test accuracies in the figure (Stratos, 2004) (12).

Based on the System column, there are 5 different parsers: McDonald13 (a cross-lingual parser) (Mcdonald et al., 2013), Yara (beam 64) (a beam-search transition-based parser) (Rasooli and Tetreault, 2015), which is based on the rich non-local features in Zhang and Nivre's study (Zhang and Nivre, 2011), K&G16 (a basic BiLSTM parser) (Kiperwasser and Goldberg, 2016), Dyer15 (a greedy word-level transition-based parser based on Stack LSTM) (Dyer et al., 2015), and Ballesteros15 (a greedy character-level transition-based parser based on Stack LSTM) (Ballesteros et al., 2015), that Stratos compared with his parser, KoreanNet (a BiLSTM parser of Kiperwasser and Goldberg's study in 2016) (Kiperwasser and Goldberg, 2016), plugged in with a jamo architecture that uses the DyNet library (Neubig et al., 2017).

Based on the Features column, the KoreanNet parser have 4 different iterations of features (one with char, one with jamo, one with char and jamo, and one with all three, word, char, and jamo). The KoreanNet parser uniquely uses the jamo features as compared with the other 5 parsers.

| System | Features | Feature Representation | Emb | POS | Dev | | Test | |
|---|---|---|---|---|---|---|---|---|
| | | | | | UAS | LAS | UAS | LAS |
| McDonald13 | cross-lingual features | large sparse matrix | – | PRED | – | – | 71.22 | 55.85 |
| Yara (beam 64) | features in Z&N11 | large sparse matrix | – | PRED | 76.31 | 62.83 | 91.19 | 85.17 |
| | | | | GOLD | 79.08 | 68.85 | 92.93 | 88.61 |
| K&G16 | word | 31060 × 100 matrix | – | – | 68.87 | 48.25 | 88.61 | 78.95 |
| | | 298115 × 100 matrix | YES | | 76.30 | 60.88 | 90.00 | 82.77 |
| Dyer15 | word, transition | 31067 × 100 matrix | – | – | 69.40 | 48.46 | 88.41 | 78.22 |
| | | 298122 × 100 matrix | YES | | 75.99 | 59.38 | 90.73 | 83.89 |
| Ballesteros15 | char, transition | 1779 × 100 matrix | – | – | 84.22 | 76.41 | 91.27 | 86.25 |
| KoreanNet | char | 1772 × 100 matrix | – | – | 84.76 | 76.95 | 94.75 | 90.81 |
| | | 1772 × 200 matrix | | | 84.83 | 77.29 | 94.55 | 91.04 |
| | jamo | 500 × 100 matrix | – | – | 84.27 | 76.07 | 94.59 | 90.77 |
| | | 500 × 200 matrix | | | 84.68 | 77.27 | 94.86 | 91.46 |
| | char, jamo | 2272 × 100 matrix | – | – | 85.35 | 78.18 | 94.79 | 91.19 |
| | | 2272 × 200 matrix | | | 85.74 | 78.76 | 94.55 | 91.31 |
| | word, char, jamo | 302339 × 200 matrix | YES | – | **86.39** | **79.68** | **95.17** | **92.31** |

Figure 5: Results comparison

Based on the Feature Representation column, the matrices correspond with the feature embeddings for each parser based on the features used in the Features column. For instance, in jamo-based KoreanNet parser, Stratos used 500 x 100 and 500 x 200 jamo embeddings and in the K&G16 word-based parser, Kiperwasser and Goldberg used 31060 x 100 and 298115 x 100 word embeddings.

The Emb column lists whether or not the parser uses pre-trained feature embeddings based on the features listed in the Features column. For instance, the K&G16 parser has two iterations, one iteration using pre-trained word embeddings, and one iteration without pre-trained word embeddings (- means without pre-trained feature embeddings).

The POS column lists the types of POS tag implementations that the parser uses, such as predicted (PRED) POS tags and gold (GOLD) POS tags (- means no pos tag implementation used).

The Dev column lists the development accuracy of the parsers with UAS (unlabeled attachment score) and LAS (labeled attachment score) evaluations.

The Test column lists the test accuracy of the parsers with UAS (unlabeled attachment score) and LAS (labeled attachment score) evaluations.

UAS and LAS are evaluation metrics that measure the percentage of words that have been assigned the correct head (with or without taking the dependency label into account).

### 2.3 Hidden Markov Model for POS Disambiguation

Apart from the parsing tree perspective, another study tried to address the disadvantages of using a dependency grammar that is similarly used in English, by using a Hidden Markov Model for POS disambiguation. For instance, there were Hidden Markov Models that were based on eojeols. A eojeol is a morpheme in the Korean language that identifies a independent fragment of a Korean sentence, such as 음식을 먹고 ("eating food"). The eojeol-based tagging model calculates lexical and transition probabilities with eojeols as a unit. However, one disadvantage behind this method is that it suffers from severe data sparseness problems since a single eojeol consists of many different morphemes.

As a result, morpheme-based Hidden Markov Model tagging was tried. These models assign a single tag to a morpheme regardless of the space in a sentence. While, this method of modelling had an advantage of Morpheme-based tagging, reducing data sparseness problems, it still had a disadvantage of incurring multiple observation sequences in Viterbi decoding since an eojeol can be segmented in many different ways.

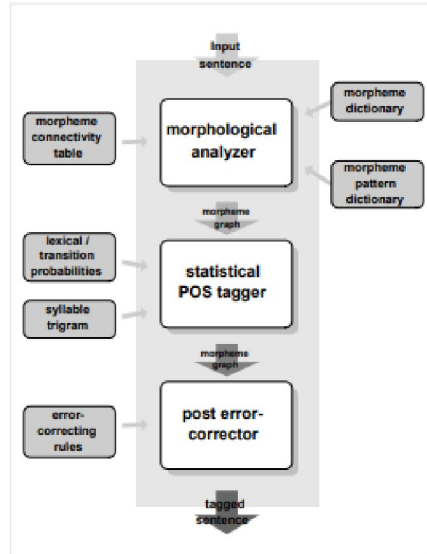

Figure 6: Statistical and rule-based hybrid architecture for POS tagging in Korean

Researchers then later tried various ways of reducing computation due to multiple observation sequences, such as shared word sequences and virtual words and two-ply HMM for morpheme unit computation but restricted within a eojeol (Song et al, 2019) (10). However, since statistical approaches take neighboring tags into account only within a limited window (usually two or three), sometimes the decision fails to cover important linguistic contexts necessary for POS disambiguation. Also, approaches using only statistical methods are inappropriate for idiomatic expressions, for which lexical terms need to be directly referenced. And especially, statistical approaches alone do not suffice for agglutinative languages, which usually have complex morphological structures (Lee et al, 2000) (6).

In agglutinative languages, a word usually consists of one or more stem morphemes plus a series of functional morphemes; therefore, each morpheme should receive a POS tag appropriate to its functional role to cope with the complex morphological phenomena in such languages.

As a result, one study proposed a hybrid method that combines statistical and rule-based approaches to POS disambiguation and can be tightly coupled with generalized unknown-morpheme-guessing techniques shown in Figure 3 (Lee et al, 2002) (5).

Before we go over a brief description of the hybrid model, we have to note that the authors used a morpheme dictionary that helps the model to identify morphemes more accurately. They use this dictionary during their tasks of morphological analysis before using the hybrid model, which involve, morpheme segmentation, recovering original morphemes from spelling changes, and morphotactic modeling.

For morpheme segmentation, the input texts are scanned from left to right, character by character, to be matched with morphemes in the morpheme dictionary. The morpheme dictionary has a trie structured index for fast matching, and has an independent entry for each variant surface form of the original morpheme so the original morphemes can easily be reconstructed from spelling changes.

For morphotactic modeling, they used the POS tags and the morphotactic adjacency symbols in the morpheme dictionary. The POS tags provide information about grammatical connectivity between morphemes needed to form an eojeol. The full hierarchy of POS tags and morphotactic adjacency symbols is encoded in the morpheme dictionary for each morpheme.

For unknown morpheme segmentation, they used the morpheme dictionary so that the system can look up unknown morphemes exactly the same way it looks up known registered morphemes. Once the unknown morphemes are identified and recovered using the pattern dictionary, the system can use the same information about adjacent morphemes in the unknown morphemes' eojeol that it would use if they were known morphemes.

After the morpheme analysis is finished, then they ran through the statistical and rule-based hybrid tagging model. Based on Figure 3, the hybrid tagging model consists of three parts: the morphological analyzer with unknown-morpheme handler, the statistical POS tagger, and the rule-based error corrector.

For the morphological analyzer with unknown-morpheme handler, the analyzer segments the morphemes from input eojeols and reconstructs the original morphemes from spelling changes by recovering the irregular conjugations. And the unknown-morpheme handler assigns initial POS tags to morphemes that are not registered in the morpheme dictionary.

For the statistical POS tagger, the tagger runs the Viterbi algorithm (a well known algorithm discussed by Forney, 1973) on the morpheme graph to search for the optimal tag sequence for POS disambiguation.

For the rule-based error corrector, the corrector is a transformer (a well known architecture discussed by Brill 1995) that corrects mistagged morphemes by consulting lexical patterns and necessary contextual information.

**Table 8**
Performance of the statistical tagger (all in %) on three document sets, using three progressively degraded versions of the tagger.

| Document set | Version 1 | Version 2 | Version 3 |
|---|---|---|---|
| Set 1 | 96.4 | 89.5 | 87.1 |
| Set 2 | 96.0 | 92.8 | 89.0 |
| Set 3 | 96.7 | 88.7 | 84.8 |
| Total | 96.4 | 90.3 | 87.0 |

Figure 7: Statistical Tagger Performance

For the morphological analysis, they used a 130,000-morpheme balanced training corpus for statistical parameter estimation, and a 50,000-morpheme corpus for learning the a posteriori error correction rules. The training corpus was collected from various sources such as Internet documents, encyclopedias, newspapers, and school textbooks.

For the test sets, they carefully selected three different document sets. The first document set of 25,299 morphemes and 1,338 sentences, which was collected from the Kemong Encyclopedia, a hotel reservation dialogue corpus, and assorted Internet documents, containing about 10% unknown morphemes. The second document set of 15,250 morphemes and 574 sentences consists solely of Internet documents from assorted domains, such as broadcasting scripts and newspapers, containing about 8.5% unknown morphemes. The third document set of 20,919 morphemes and 555 sentences consists of a standard Korean document set called KTSET 2.0 including academic articles and electronic newspapers, containing about 14% unknown morphemes (mainly technical jargon).

As Figure 4 shows, the performance of the statistical tagging part of the hybrid model is shown with these three sets of documents with 3 progressively degraded versions of the tagging mechanism. Version 1 is a full version using the statistical method. Version 2 is a somewhat degraded version that does not use the system's unknown-morpheme guessing capability but treats all the segmented unknown morphemes as nouns. Version 3 is an even more degraded version that rejects all unknown-morphemes as tagging failures (this version does not even perform unknown-morpheme segmentation during morphological analysis. This experiment with the statistical tagging verifies the effectiveness of their unknown-morpheme segmentation and guessing techniques, as shown by the sharp performance drops between Versions 1, 2, and 3.

**Table 9**
Performance of the hybrid tagger (all in %) on three document sets, using two versions of the tagger.

| Document set | Version 1 | Version 2 |
|---|---|---|
| Set 1 | 97.2 | 96.4 |
| Set 2 | 96.9 | 96.0 |
| Set 3 | 97.4 | 96.7 |
| Total | 97.2 | 96.4 |

Figure 8: Hybrid Tagger Performance

As Figure 5 shows, the performance of the hybrid tagger were tested with the same three document sets. Version 1 was the full POSTAG system with unknown-morpheme segmentation, guessing, and rule-based error correction. Version 2 did not employ a posteriori error correction rules (the same syste as Version 1 in the statistical tagging in Figure 4). The performance dropped between Version 1 and Version 2, however the drop rates were mild due to the performance saturation at Version 1, which means that the statistical tagger alone already achieves the best performance for tagging Korean morphemes.

Although the authors didn't compare this hybrid based model directly with previous models, they have shown that their model is effective through their high recall and precision scores of various POS tags, which is at the state-of-the-art level of performance of POS tagging of unknown morphemes.

**2.4 Conclusions**

From each of these three different methods of parsing through Korean, each have advantages and disadvantages of their own. The first method, proposed by Chung et al., used a probabilistic approach to statistical parsing. This method had some advantages that include that it performs really well with the training data since it memorizes the lexical dependency relation when learning and applies it untouched in testing. However, the disadvantages arise when the model deals with unlexicalized data, which means that the memorizing of the dependencies won't be as effective. For the method proposed by Stratos, it used a Bi-LSTM in order to parse through jamo letters in Korean characters of Korean sentences. The advantages lie in the feature selection as opposed to the morpheme fragments of Chung's model, which resulted in a higher accuracy. However, the disadvantages still arise with the sparseness of data, which still arise since there are unknown morphemes to consider. For the method proposed by Gary et al., the advantages lie in the morphological analysis method that was being used in conjunction with the hybrid hidden markov model, which is applied with rule-based error correction. However, some possible disadvantages arise when there are unknown morphemes, which would lead to unknown morpheme estimation rather than more accurate dictionary lookups. It is clear that the Hybrid Hidden Markov Model for POS Disambiguation that was proposed by Gary et al., 2002, as shown the highest accuracy out of all three aforementioned methods.

# 3 Chinese (Mandarin)

Chinese consists of hanzi characters. Each character represents a syllable and each word usually consists of one or more character tokens. Due to the lack of word delimiters (e.g., white space) in Chinese text, similar to Japanese, Chinese text segmentation is more difficult than English text segmentation.

Chinese word segmentation (CWS) is challenging partly because it is difficult to define what constitutes a word in Chinese (Gao et al, 2005) (3). Due to the particularity of Chinese, the two major difficulties involve ambiguity and out of vocabulary. For example, in terms of ambiguity, the right segmentation of "部分居民生活水平"(some residents' living standard) should be "部分/居民/生活/水平"(some / residents'/ living / standard). However, "分居" (separation) and "民生"(people's Livelihood) can also form correct words. The second issue is out of vocabulary. Those words are not included in the training set but encountered in the test data, such as, terminology in scientific fields like economics, medical treatment, and technology, or new words on social media, or people's name. Such problems are especially obvious in cross-domain word segmentation tasks. Approaches have been changing all the time.

## 3.1 Character-based Approaches

Back in 2005, character-based approaches were dominant solutions for CWS. People usually define CWS as a character-based sequence labeling task and use methods including Conditional Random Field (Tseng et al, 2005) (14), Maximum Entropy Modeling (Song et al, 2006) (11) and so on.

### 3.1.1 Conditional Random Field

Conditional Random Field (CRF) is a statistical sequence modeling framework similar to Maximum Entropy Markov Model (MEMM) and this discriminant framework treats the task as a binary decision task with each character being labeled either as a word's beginning or otherwise. It used Gaussian priors to prevent overfitting and a quasi-Newton method to optimize parameters. One of the advantages I've observed was that under the help of CRF, we can utilize a large amount of n-gram features and different state sequence-based features. CRF also allows us to have a framework for the use of morphological features. One disadvantage of this method is that it has a possibility of having its performance to drop when adding unknown word features to the HK corpus. However, the authors state that the testing data uses different punctuation than the training set, meaning that their system could not distinguish new word characters from new punctuation. If the new punctuation was not unknown, then the performance of the data of the HK corpus would increase, with the unknown word features to have neglible effect to the performance.

### 3.1.2 Maximum Entropy Modeling

Maximum Entropy Modeling (MEM) is another method in terms of character-based sequence labeling. It is fully automatically generated from the training data set by training and analyzing the character sequence from the given training corpus. In Song et al's paper (Song et al, 2006) (11), they shifted the idea from focusing on the dealing with the character itself to getting the segment probability and where to split the character sequence. This model enables generation of diverse and accurate segmentation lattices from a single model that are appropriate for use in decoders that accept word lattices as input. Similarly with the Conditional Random Field method, MEM's main disadvantage is due to the fact that it is based on a closed track system, meaning that it cannot handle unknown characters. In terms of MEM, there is a possibility that the sentence being segmented might have a character written that does not appear in the training corpus, which would lead to a decrease in performance by the model. However, the advantage with this model is that it is based on a neural network model and that it incorporates contextualized information during segmentation.

In summary, all these various character-based sequence labeling methods are used to effectively extract contextual features and to help better predicting segmentation labels for each character.

## 3.2 Neural Network-based Approach

With the advent of the Neural Network, recent approaches incorporating NN techniques are proved to be powerful in dealing with CWS. However, the NN architecture can provide a more accurate model but not explicit wordhood information. One outstanding advantage of non-neural models is that explicitly extracting n-grams can provide more wordhood information of the context.

### 3.2.1 Wordhood Memory Module

A recent paper by Tian et al (Tian et al, 2020) (13) proposed an improved neural CWS approach combined with Wordhood Memory Module called WMSEG. The detailed architecture is illustrated in Figure 5.

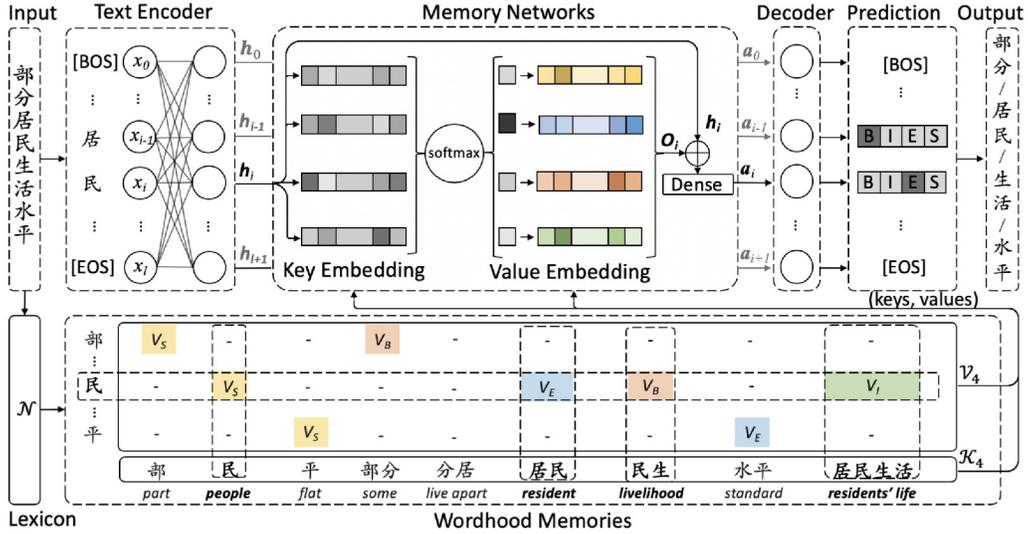

Figure 9: The architecture of WMSEG

The input sentence is "部分居民生活水平" (some residents' living standard) and the output segmented sentence is "部分/居民/生活/水平" (some / residents' / living / standard). The general sequence labeling paradigm is the top part and this model has an extra memory module inserted between the Text Encoder and the Decoder. First, from the given input sentence, we extract n-grams and then get the wordhood information of those n-grams from the Lexicon. Passing the output from the Text encoder, n-grams, and wordhood information to the Memory Networks, and then passing the output into a decoder, we can get the final predictions. The model predicts a tag for each character and then converts it to word boundary as output.

The innovative part of this model is the Wordhood Memory Networks. Compared to other structures that exploit n-grams such as the attention mechanism, this key-values based memory network is more efficient to model pairwise knowledge and offers valuable information between segmented contexts and their meanings. This model also introduces flexibility to integrate various prevailing encoders and decoders to further improve the performance. The encoders used in the paper are Bi-LSTM, BERT, and ZEN and the decoders are softmax and CRF. Here's the experimental results of WMSEG on SIGHAN2005 and CTB6, which are two well-known Chinese Language Processing datasets.

In Figure 6, "ENDN" stands for the text encoders and "WM" is whether the wordhood memories are used or not. "BL" is Bi-LSTM and "BT" is BERT. "SM" is softmax and "CRF" is CRF. The bold numbers indicate the higher scores. We can clearly see that the overall performance is improved when adopting the wordhood memory module (MSR, PKU, AS, CITYU, and CTB6 are 5 benchmark datasets used for the experiment, the $R_{OOV}$ is the R value for out of vocabulary words, and F is the F measure).

| Config | | MSR | | PKU | | AS | | CityU | | CTB6 | |
|---|---|---|---|---|---|---|---|---|---|---|---|
| En-Dn | WM | F | $R_{OOV}$ | F | $R_{OOV}$ | F | $R_{OOV}$ | F | $R_{OOV}$ | F | $R_{OOV}$ |
| BL-SM | × | 95.53 | 62.96 | 91.85 | 48.84 | 94.52 | 62.21 | 93.79 | 67.26 | 93.56 | 67.39 |
|  | ✓ | **95.61** | **63.94** | **91.97** | **49.00** | **94.70** | **64.18** | **93.88** | **69.20** | **93.70** | **68.52** |
| BL-CRF | × | 95.80 | 66.17 | 92.35 | 52.04 | 94.39 | 61.59 | 93.96 | 67.84 | 93.84 | 70.81 |
|  | ✓ | **95.98** | **68.75** | **92.43** | **56.80** | **95.07** | **68.17** | **94.20** | **69.91** | **94.03** | **71.88** |
| BT-SM | × | 97.84 | 86.32 | 96.20 | 84.43 | 96.33 | 77.86 | 97.51 | **86.69** | 96.90 | **88.46** |
|  | ✓ | **98.16** | **86.50** | **96.47** | **86.34** | **96.52** | **78.67** | **97.77** | 86.62 | **97.13** | 88.30 |
| BT-CRF | × | 97.98 | 85.52 | 96.32 | 85.04 | 96.34 | 77.75 | 97.63 | 86.66 | 96.98 | 87.43 |
|  | ✓ | **98.28** | **86.67** | **96.51** | **86.76** | **96.58** | **78.48** | **97.80** | **87.57** | **97.16** | **88.00** |
| ZEN-SM | × | 98.35 | **85.78** | 96.27 | 84.50 | 96.38 | 77.62 | 97.78 | 90.69 | 97.08 | 86.20 |
|  | ✓ | **98.36** | 85.30 | **96.49** | **84.95** | **96.55** | **78.02** | **97.86** | **90.89** | **97.22** | **86.83** |
| ZEN-CRF | × | 98.36 | **86.82** | 96.36 | 84.81 | 96.39 | 77.81 | 97.81 | **91.78** | 97.13 | 87.08 |
|  | ✓ | **98.40** | 84.87 | **96.53** | **85.36** | **96.62** | **79.64** | **97.93** | 90.15 | **97.25** | **88.46** |

Figure 10: Experimental results of WMSEG on SIGHAN2005 and CTB6 datasets with different configurations

In summary, WMSEG proposed a CWS approach based on key-value memory neural network. WMSEG maps n-grams and their wordhood information to corresponding keys and values. It also appropriately models the values according to the importance of keys in a specific context, which can decrease the effect of ambiguity. The vocabulary is constructed through an unsupervised method to realize the use of unlabeled text in a specific field, thereby improving the recognition of unseen words. Out of the three methods, CRF, MEM, and WMSEG, WMSEG yields the best accuracy.

# 4 Japanese

Japanese, in contrast to Chinese, includes some character types like hiragana, katakana, and kanji. They produce orthographic variations and thus make the word segmentation process more challenging (Choi et al, 2009) (1).

## 4.1 Hidden Markov Model

In their decision-making method, some early Japanese word segmentation techniques rely consistently on broad lexicons of words. A standard word processor would have both a knowledge base that encodes Japanese grammar rules and more than 50,000 word lexicons. Hypothesized segments of the incoming text are evaluated to decide whether they have any semantics and are thus more likely to be correct segments; grammatical rules are then applied to make final segmentation decisions. For example, Dragon Systems' LINGSTAT machine translation method uses a maximum probability segmentation algorithm that basically measures all possible segmentations of a sentence using a large lexicon and chooses the one with the best score or probability (Papageorgiou et al, 1994) (8).

In Papageorgiou's work, the Hidden Markov Model (HMM), which is a stochastic algorithm, is used to determine the boundary of the words. One potential way to segment is to use the Hidden Markov model to find the most probable word sequence based on the brute force calculation of any word sequence. This, though, breaks the restriction of not depending on the Japanese word lexicon.

Given that we would prefer to avoid the overhead involved with forming and using a word-based lexicon, we are thus compelled to address the issue in a way that reflects on discrete characters and their interrelationships. The Hidden Markov Model itself is able to achieve 91 percent accuracy.

## 4.2 Hybrid Method with Hidden Markov Model and char-level information

In Nakagawa's work, a hybrid method of Hidden Markov Model (HMM) and word-level and character-level information is used. One of the problems with word segmentation is that the word segmentation algorithm does not know how to handle out of vocabulary words, and thus the task of determining word boundaries for such words is difficult (Nakagawa et al, 2000) (7). The Hidden Markov model-based approach is found to have a high overall accuracy, but the accuracy decreases for unknown terms, thus character level information is used in addition to HMM to improve the accuracy. The Character Tagging Method refers to a way to separate the sentence by labelling words. There are 4 main types of tags: B(beginning of a word), I(middle of a word), E(end of a word), S(itself is a word). In this way, a sentence like 昨日□校に行きました can be marked as 昨(B)日(E) | □(B)校(E) | に(S) | 行(B)き(E) | た(S). Unknown words can easily be dealt using this labeling system because known words and unknown words are treated similarly and no other extraordinary processing is needed.

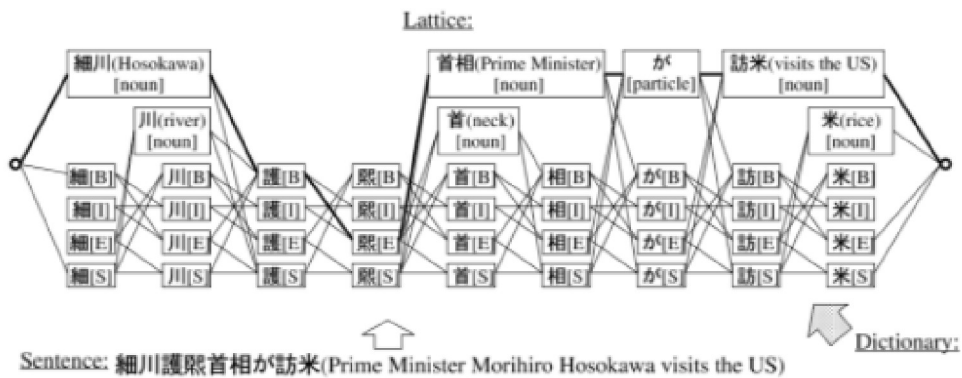

Figure 11: Word Lattice for this Japanese sentence used for HMM

The hybrid approach is primarily based on word-level Markov models, but both POC-tags and POS-tags are used concurrently, and word segmentation for known words and unknown words is done simultaneously. The following diagram(word lattice) is an example. Here, the second and third characters are the name of a person, which is an unknown word for the system.

The probability of the unknowing term is approximated by the variable character probability product, and the calculation is performed in the sense of the Markov model-based word-level method.

### 4.3 LSTM with word dictionary and n-gram embedding

Besides Hidden Markov Models, JWS can also be done in a neural network way. Traditional machine learning methods usually will have data sparseness problems and lack global information in the sentences. And Neural network methods are able to address them (Kitagawa, 2017) (4).

Kitagawa and Komachi proposed a LSTM network for Japanese word segmentation (JWS). Same as the HMM described above, they are also using the character level embedding to optimize the algorithms. The detailed architecture is shown in Figure 12. Recurrent neural network (RNN) models can grasp long distance information owing to the use of long short– term memory (LSTM), achieving state-of-the-art accuracy. In addition to this, Character-Type Embeddings, Character-Based n-gram Embeddings, and word dictionary are incorporated into the LSTM as well to optimize the performance. According to the training result, the combination of LSTM, n-gram language model and dictionary is able to achieve the highest F1 score of 98.42.

### 4.4 Conclusions

In conclusion, the pros and cons of each methods are very obvious. Papageorgiou's Hidden Markov Model is easy to understand and does not need a lot of cost to train. But its ability to determine the boundary of words is limited and thus can only achieve a low accuracy. Nakagawa's hybrid model, which is made by both Hidden Markov Model and word level models. This takes more effort but gives the algorithms the ability to handle out-of-vocabulary words, as word-level and char-level information will enable the algorithm to guess the Part-Of-Speech Tags. However, traditional JWS approaches still rely on local features. This means even though these traditional machine learning approaches can work properly with the help of language models, they still have the data sparseness problem. This means they require much more data to train than deep learning algorithms like neural networks do. In Kitagawa and Komachi's work, they employed an LSTM-based approach to JWS. The results proved that it is the most effective method of the three methods covered above.

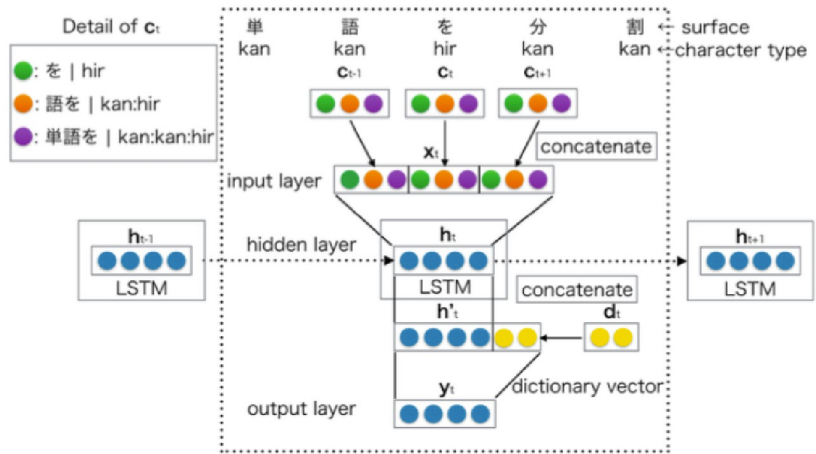

Figure 12: LSTM architecture

## 5 Related work

As mentioned previously in the introduction, there has been work in trying to bridge the various different languages with different characteristics together with a universal dependency. In order to tackle the issue with having different languages, with different characteristics to consider when segmenting them, the authors listed the typological factors that depend more on writing system, shown as follows: Character Set Size, Lexicon Size, Average Word Length, Segmentation Frequency, Multiword Token Portion, and Multiword Token Set Size. In general, the authors have used different methods of segmentation such as the Bi-RNN CRF model, and transduction and considered language specific characteristics such as character representation, space delimiters, and non-segmented multi-word tokens (which work well with Arabic and Hebrew languages). In general, as a form of extrinsic evaluation, the authors test the segmenter in a dependency parsing setup on the datasets where they obtained substantial improvements over UDPipe. From their experiment, we can see that word segmentation accuracy has a great impact on parsing accuracy as the segmentation errors propagate. From the results, which included the results in table 9, we concluded that having a more accurate word segmentation model is very beneficial for achieving higher parsing accuracy. The main reason why this work is related to our survey is that this work expands not only for Asian Languages, but to a significantly larger amount of languages from different global regions, such as western and eastern Europe, South America, the Middle East, and etc. (In table 9, UDPipe is another segmentation model developed by Straka and Strakova, 2017, that contains word segmentation, POS tagging, morphological analysis and dependency parsing models in a pipeline) (9).

| Dataset | UDPipe | This Paper | Dataset | UDPipe | This Paper | Dataset | UDPipe | This Paper |
|---|---|---|---|---|---|---|---|---|
| Ancient Greek | 99.98 | 99.96 | Ancient Greek-PROIEL | 99.99 | 100.0 | Arabic | 93.77 | 97.16 |
| Arabic-PUD | 90.92 | 95.93 | Basque | 99.97 | 100.0 | Bulgarian | 99.96 | 99.93 |
| Catalan | 99.98 | 99.80 | Chinese | 90.47 | 93.82 | Croatian | 99.88 | 99.95 |
| Czech | 99.94 | 99.97 | Czech-CAC | 99.96 | 99.93 | Czech-CLTT | 99.58 | 99.64 |
| Czech-PUD | 99.34 | 99.62 | Danish | 99.83 | 100.0 | Dutch | 99.84 | 99.92 |
| Dutch-LassySmall | 99.91 | 99.96 | English | 99.05 | 99.13 | English-LinES | 99.90 | 99.95 |
| English-PUD | 99.69 | 99.71 | English-ParTUT | 99.60 | 99.51 | Estonian | 99.90 | 99.88 |
| Finnish | 99.57 | 99.74 | Finnish-FTB | 99.95 | 99.99 | Finnish-PUD | 99.64 | 99.39 |
| French | 98.81 | 99.39 | French-PUD | 98.84 | 97.23 | French-ParTUT | 98.97 | 99.32 |
| French-Sequoia | 99.11 | 99.48 | Galician | 99.94 | 99.97 | Galician-TreeGal | 98.66 | 98.07 |
| German | 99.58 | 99.64 | German-PUD | 97.94 | 97.74 | Gothic | 100.0 | 100.0 |
| Greek | 99.94 | 99.86 | Hebrew | 85.16 | 91.01 | Hindi | 100.0 | 100.0 |
| Hindi-PUD | 98.26 | 98.82 | Hungarian | 99.79 | 99.93 | Indonesian | 100.0 | 100.0 |
| Irish | 99.38 | 99.85 | Italian | 99.83 | 99.54 | Italian-PUD | 99.21 | 98.78 |
| Japanese | 92.03 | 93.77 | Japanese-PUD | 93.67 | 94.17 | Kazakh | 94.17 | 94.21 |
| Korean | 99.73 | 99.95 | Latin | 99.99 | 100.0 | Latin-ITTB | 99.94 | 100.0 |
| Latin-PROIEL | 99.90 | 100.0 | Latvian | 99.16 | 99.56 | Norwegian-Bokmaal | 99.83 | 99.89 |
| Norwegian-Nynorsk | 99.91 | 99.97 | Old Church Slavonic | 100.0 | 100.0 | Persian | 99.65 | 99.62 |
| Polish | 99.90 | 99.93 | Portuguese | 99.59 | 99.10 | Portuguese-BR | 99.85 | 99.52 |
| Portuguese-PUD | 99.40 | 98.98 | Romanian | 99.68 | 99.74 | Russian | 99.66 | 99.96 |
| Russian-PUD | 97.09 | 97.28 | Russian-SynTagRus | 99.64 | 99.65 | Slovak | 100.0 | 99.98 |
| Slovenian | 99.93 | 100.0 | Slovenian-SST | 99.91 | 100.0 | Spanish | 99.75 | 99.85 |
| Spanish-AnCora | 99.94 | 99.93 | Spanish-PUD | 99.44 | 99.39 | Swedish | 99.79 | 99.97 |
| Swedish-LinES | 99.93 | 99.98 | Swedish-PUD | 98.36 | 99.26 | Turkish | 98.09 | 97.85 |
| Turkish-PUD | 96.99 | 96.68 | Ukrainian | 99.81 | 99.76 | Urdu | 100.0 | 100.0 |
| Uyghur | 99.85 | 97.86 | Vietnamese | 85.53 | 87.79 | **Average** | **98.63** | **98.90** |

Table 9: Evaluation results on the UD test sets in F1-scores. The datasets are represented in the corresponding treebank codes. *PUD* suffix indicates the parallel test data. Two shades of green/red are used for better visualisation, with brighter colours for larger differences. Green represents that our system is better than UDPipe and red is used otherwise.

## 6  Conclusions and future work

In conclusion, the word segmentation methods for Chinese, Korean, and Japanese share a lot of similarities in the early stages: statistical language models are widely applied, since determining word boundaries using probabilities generated by machine learning models is very intuitive. Neural network methods are also popular. It is shown that a combination of deep learning methods(RNN, LSTM) and character-level information embedding is able to achieve the highest accuracy. It is also able to handle problems related to unknown words.

CKJ word segmentation methods have also been applied to English processing, mainly in handwriting recognition. Because when recognizing handwriting, the spaces between words are not very clear. The Chinese word segmentation method can help distinguish the boundaries of English words. In fact, many mathematical methods of language processing are universal and have nothing to do with specific languages. When people design language processing algorithms, we always consider whether it can be easily applied to various natural languages. In this way, we can effectively support searches in hundreds of languages.

Some potential future work includes exploring or discovering newer techniques in handling an even broader array of Asian languages that haven't been segmented before with word segmentation and to use the findings from the Shao et al., 2018 study and to further analyze the advantages and disadvantages of this new method and compare the new method with other methods, perhaps the methods that were discussed in this survey of the CJK langauges.

## 7  Team Responsibility

Matthew was responsible for researching about word segmentation models in the Korean language and researching about related work in the topic. Qin was responsible for researching about word segmentation models in the Japanese language and future work. Yexin was responsible for researching about word segmentation models in the Chinese language and the significance of word segmentation in the field of natural language processing.